\documentclass[journal,twoside,web]{ieeecolor}
\usepackage{generic}
\usepackage{cite}
\usepackage{amsmath,amssymb,amsfonts}
\usepackage{algorithmic}
\usepackage{graphicx}
\usepackage{algorithm,algorithmic}
\usepackage{hyperref}
\usepackage{booktabs}
\hypersetup{hidelinks}
\usepackage{textcomp}
\usepackage{threeparttable}
\def\BibTeX{{\rm B\kern-.05em{\sc i\kern-.025em b}\kern-.08em
    T\kern-.1667em\lower.7ex\hbox{E}\kern-.125emX}}
\begin{document}
\title{Shared Attention-based Autoencoder with Hierarchical Fusion-based Graph Convolution Network for sEEG SOZ Identification}
\author{Huachao Yan, Kailing Guo*, \IEEEmembership{Member, IEEE}, Shiwei Song*, Yihai Dai, Xiaoqiang Wei, Xiaofen Xing, \IEEEmembership{Member, IEEE} and Xiangmin Xu, \IEEEmembership{Senior Member, IEEE}.
\thanks{This work is supported in part by the Fundamental Research Funds for Central Universities, SCUT, under Grant 2023ZYGXZR086, in part by the National Key R\&D Program of China under Grant 2022YFB4500600, in part by the National Natural Science Foundation of China under Grant 61802131, in part by the Science and Technology Project of Guangdong under Grant 2022B0101010003, in part by the Startup Fund for Scientific Research of Fujian Medical University under Grant 2021QH1048, in part by the Fujian Provincial Natural Science Foundation Program under Grant 2021J01788, in part by the Guangzhou Key Laboratory of Body Data Science under Grant 201605030011, in part by the Science and Technology Project of Zhongshan under Grant 2019AG024, in part by the Guangdong Provincial Key Laboratoryof Human Digital Twin under Grant 2022B1212010004.}
\thanks{\noindent Huachao Yan,  Kailing Guo and Xiaofen Xing are with the School of Electronic and Information Engineering, South China University of Technology, Guangzhou 510641, China.}
\thanks{\noindent Yihai Daiand Xiaoqiang Wei are with the Department of Neurosurgery, Fujian Medical University Union Hospital, Fuzhou, 350001, Fujian, China.}
\thanks{\noindent Shiwei Song is with the Department of Neurosurgery, Fujian Medical University Union Hospital, Fuzhou, 350001, Fujian, China; and the Department of Pediatric Epilepsy, Fuzhou Children’s Hospital, Fuzhou, 350005, Fujian, China}
\thanks{\noindent Xiangmin Xu is with the School of Electronic and Information Engineering, South China University of Technology, Guangzhou 510641, China; Pazhou Lab, Guangdong Artifical Intelligence and Digital Economy Laboratory, Guangzhou, 510330, Guangdong, China; and the Institute of Aritificial Intelligence, Hefei Comprehensive National Science Center, Heifei, 230088, Anhui, China}
\thanks{*Corresponding authors}
}
%

\maketitle

\begin{abstract}
Diagnosing seizure onset zone (SOZ) is a challenge in neurosurgery, where stereoelectroencephalography (sEEG) serves as a critical technique.
In sEEG SOZ identification, the existing studies focus solely on the intra-patient representation of epileptic information, overlooking the general features of epilepsy across patients and feature interdependencies between feature elements in each contact site.
In order to address the aforementioned challenges, we propose the shared attention-based autoencoder (sATAE). 
sATAE is trained by sEEG data across all patients, with attention blocks introduced to enhance the representation of interdependencies between feature elements. 
Considering the spatial diversity of sEEG across patients, we introduce graph-based method for identification SOZ of each patient. 
However, the current graph-based methods for sEEG SOZ identification rely exclusively on static graphs to model epileptic networks.
Inspired by the finding of neuroscience that epileptic network is intricately characterized by the interplay of sophisticated equilibrium between fluctuating and stable states, we design the hierarchical fusion-based graph convolution network (HFGCN) to identify the SOZ.
HFGCN integrates the dynamic and static characteristics of epileptic networks through hierarchical weighting across different hierarchies, facilitating a more comprehensive learning of epileptic features and enriching node information for sEEG SOZ identification.
Combining sATAE and HFGCN, we perform comprehensive experiments with sATAE-HFGCN on the self-build sEEG dataset, which includes sEEG data from 17 patients with temporal lobe epilepsy.
The results show that our method, sATAE-HFGCN, achieves superior performance for identifying the SOZ of each patient, effectively addressing the aforementioned challenges, providing an efficient solution for sEEG-based SOZ identification.

\end{abstract}

\section{Introduction}
\label{sec:introduction}
\IEEEPARstart{A}{ccording} to data from the World Health Organization (WHO), the number of people with refractory epilepsy with drug resistance approaching 50 million in the world \cite{rf1}, with an annual increase of approximately 2.5 million cases a year \cite{rf2}. 
Among patients with epilepsy, 40-50\% are diagnosed with temporal lobe epilepsy.
Epilepsy can be regarded as abnormal electrical discharges originating from a specific region within the brain, known as the seizure onset zone (SOZ).

EEG (Electroencephalography) has achieved great success in the fields of brain-computer interfaces \cite{rf79} and brain disease diagnosis \cite{rf78}.
In the process of diagnosing the SOZ, two types of EEG are crucial: surface EEG and stereoEEG (sEEG).
Surface EEG is commonly employed for the initial screening and continuous monitoring of epilepsy patients, due to its convenient use and non-invasive characteristics \cite{rf81}.
However, surface EEG can only record the overall brain activity expressed on the cortical surface, making it difficult to localize SOZ in deeper brain regions.
sEEG is an invasive EEG that enables the recording of neural activity from specific deep brain regions with high precision. 
Consequently, sEEG has become a key technique for localizing SOZ \cite{rf4}.
sEEG is obtained by implanting multiple high-density electrodes, each equipped with numerous evenly distributed contact sites, to capture electrical activity at various spatial locations within the brain.
In recent years, many studies have conducted statistical analyses of sEEG across different states and found that the characteristics of sEEG vary significantly among various behavioral states, such as sleep \cite{rf5}, wakefulness \cite{rf6}, and seizures \cite{rf7}.
Additionally, sEEG with high-density electrodes also enables measuring brain connectivity through the cortical-cortical evoked potential (CCEP), which is obtained by stimulating a specific brain region electrically with contact sites on a high-density electrode.
The resulting evoked responses recorded at other contact sites, distant from the stimulation site, provide information revealing the synaptic connectivity between the stimulated and recording regions \cite{rf8}, \cite{rf9}.

Typically, commonly used preprocessing for sEEG involves  decomposing the signal of each contact site into six frequency bands ($\delta$ (1-4 Hz), $\theta$ (4-8 Hz), $\alpha$ (8-15 Hz), $\beta$ (15-30 Hz), low $\gamma$ (30-80 Hz), and high gamma $\gamma$ (80-150 Hz)) \cite{rf11} and then calculate features such as power spectra \cite{rf5} and root mean square (RMS) values \cite{rf10} from each frequency band.
With the development of deep learning, 
the work \cite{rf13} further extracts highly representative latent features from handcrafted feature using autoencoder .
The advantage of using an autoencoder lies in its ability to capture complex patterns within a large amount of data \cite{rf60}, \cite{rf61}.
However, existing study \cite{rf13} on sEEG-related autoencoder training strategy has potential for enhancement, as they predominantly focus on training separate autoencoders by using a small amount of data from each patient.
This training strategy fails to capture the general patterns of epileptic neural activity across all patients, resulting in insufficient method generalization \cite{rf15}.
In addition, relying solely on autoencoders for feature extraction leads to inadequate representation quality \cite{rf14}, as it overlooks distinct interdependencies between feature elements for epileptic information representation.

To address the above issue, this paper designs a shared attention-based autoencoder (sATAE).
We leverage sEEG data from all patients to train a single sATAE, enhancing the method's generalizability in representing epileptic information.
Furthermore, we propose the pooling-based attention block and incorporate it into the encoder of sATAE, capturing the different importance of feature elements for epileptic information and thereby improving the quality of feature representation. 

Many studies designed deep learning methods to classify the obtained sEEG features, achieving considerable results.
Among the deep learning methods for sEEG SOZ identification, convolution neural networks (CNN) and graph neural networks (GNN) have received much attention for their effectiveness. 
Some studies \cite{rf16}, \cite{rf17}, \cite{rf18} employ convolution operation on each sEEG contact site for SOZ identification.
However, the findings of neuroscience have shown that spatial collaboration exists between different brain regions \cite{rf19}, \cite{rf20}, and CNN-based methods overlook the spatial relationships between contact sites.
In addition, contact sites are located in different brain regions, exhibiting varying connectivities and connection strengths, demonstrating unstructured relationships.
Graph-based methods are able to cope with unstructured data directly and hold promise in describing the intrinsic spatial relationships between different nodes in the graph.
For sEEG SOZ identification, many studies focus on employing static relationships to construct the adjacency matrix within the epileptic network from different aspects, e.g., correlations \cite{rf22} and distance \cite{rf23}, achieving considerable results.
However, the methods proposed in above mentioned studies lack sufficient representational capacity for sEEG SOZ identification. 

In graph-based studies, many methods have been proposed from different perspectives to enhance representational ability, including node sampling, edge dropping, and skip connection.
Node sampling \cite{rf40}, \cite{rf45} methods design sampling strategies based on different properties to select a subset of nodes from the original graph and aggregate nodes within the resulting subgraph, which can be viewed as dropping nodes to improve method performance.
To preserve node features,
edge dropping \cite{rf43}, \cite{rf44} methods are proposed, which remove a percentage of edges according to some specific strategies.
Dominik et al. \cite{rf86} introduce the edge dropping to proposed GCN for EEG-based Alzheimer's recognition.
However, applying node sampling or edge dropping to graph-based sEEG SOZ identification
will lose the interpretability of brain connectivity.
Motivated by CNN, some studies \cite{rf39}, \cite{rf42} introduce the different types of skip connections into graph methods, which effectively enhance representational ability while preserving interpretability.
Gao et al. \cite{rf84} integrate the skip connection and uncertainty learning into GCN to learn uncertainty connectivity of brain network for EEG emotion recognition.
Yang et al. \cite{rf85} further consider the varying importance of different EEG channels and incorporate skip connections and channel weighting to enhance EEG-based emotion recognition.
However, the aforementioned studies use regular skip connections, which primarily focus on integrating information from a singular perspective.
Neuroscience reveals that the brain network is intricately characterized by the interplay of both dynamic and static perspectives, reflecting a sophisticated equilibrium between fluctuating and stable states \cite{rf24}.
Neuronal populations in the brain also exhibit dynamic fluctuations in neural synchronization, which are also evident in the propagation of epilepsy \cite{rf46}, \cite{rf47}.
Nonetheless, the regular skip connections lacking the capability to integrate information interactions across different hierarchical levels between these two perspectives and raise the issue of insufficient representational ability for sEEG SOZ identification.

To tackle the above issues, this paper proposes a hierarchical fusion-based graph convolution network (HFGCN) and employ it on each patient for sEEG SOZ identification. 
HFGCN leverages both dynamic and static convolutions to mode; the intricate coexistence patterns of dynamic and static components within epileptic network.
Specifically, static convolution is employed to learn high-order topological representations from the stable information of the brain network.
The outputs from each static convolution layer are then fed into dynamic graph convolution layers to capture the dynamic characteristics of the epileptic network. 
Subsequently, we integrate the results from both static and dynamic convolutions to generate corresponding weights, and use these weights to fuse the integrated output from the subsequent layers.
HFGCN effectively captures the varying importance of spatial features across different hierarchies from static and dynamic epileptic networks and comprehensively enriches node information to improve SOZ identification performance.

\section{Related Works}
\subsection{AutoEncoder}
Basically, an autoencoder is composed of an encoder and a decoder \cite{rf74}.
Both the encoder and decoder have the same structure which includes multiple fully connected layers.
Early studies \cite{rf60} primarily focus on the autoencoder's ability to unsupervisedly compress features effectively, learning efficient representations by compressing high-dimensional data into a latent space and reconstructing the original input.
With the development of training strategies,
increasing studies \cite{rf12} has highlighted the autoencoder's ability to learn common patterns from large amounts of data by extracting general features across different samples \cite{rf60}, \cite{rf61}.
Benefiting from the autoencoder's capabilities mentioned above, the autoencoder has achieved great success in image segmentation \cite{rf31} and biomedical signal process \cite{rf83}.

In sEEG SOZ identification, the current study also uses autoencoder to perform feature processing.
Dou et al. \cite{rf13} employ an autoencoder to learn low-dimensional latent representations from sEEG data across three behavioral states for each patient.
However, this study trains a separate autoencoder for each patient, limiting its ability to extract general epileptic information across individuals.
Furthermore, relying exclusively on autoencoders for feature extraction results in inadequate feature quality, as this method overlooks interdependencies between feature elements of each contact site for sEEG SOZ identification.

\subsection{CNN Learning Method and Its Application for Epileptic sEEG Analysis}
CNN have been widely used in various fields such as signal processing \cite{rf58} and computer vision \cite{rf57} due to their ability to learn highly representative features from raw data.
In image-based tasks, 2D CNN is typically employed to capture spatial features between pixels from images. 
Distinctly, in the field of signal processing, most studies commonly apply 1D CNN to extract temporal features along a single time axis in time-series data.
Specifically, in biomedical signal analysis, CNN have shown great potential in extracting meaningful patterns from complex signals such as EEG \cite{rf78}, electrocardiogram (ECG) \cite{rf71} and electromyography (EMG) \cite{rf72}.

Most studies employ convolution operation on single contact sites to extract epileptic information from the time and frequency domain.
Wang et al. \cite{rf17} employ 1D CNN to extract epileptic information from different waveforms of sEEG in the time domain (spikes, ripples, and fast ripples)  for sEEG SOZ identification.
Graham et al. \cite{rf16} consider the characteristics of sEEG at different time scales in the time domain, and employ the ResNet 1D convolution with different kernel sizes for detection of sEEG epileptic activity.
Xiao et al. \cite{rf18} transform the sEEG signal into a spectrogram image and employ the 2D CNN to detect epileptic activity from the time-frequency domain.
Wang et al. \cite{rf82} further consider spatial relationships among sEEG channels and employ a combination of 1D and 2D CNN to capture both temporal information within sEEG and spatial information across sEEG for enhanced epileptic seizure prediction.
Although CNN-based methods can effectively extract internal features from different domains of sEEG contact sites,
neuroscience findings \cite{rf21} suggest that brain regions exhibit nonlinear spatial cooperative relationships,
CNN-based methods overlook this spatial characteristic.

\subsection{Graph Learning Method and Its Application for Epileptic sEEG Analysis}

GNN have demonstrated promising performance in managing data with spatial relationships, as seen in applications such as social network a
nalysis \cite{rf31} and traffic forecasting \cite{rf32}. With high spatial resolution and numerous recording sites, sEEG captures intricate spatial connectivity patterns, which GNN is well-suited to model and analyze effectively.
In epileptic sEEG analysis, GNN can reveal the intrinsic topological relationship among sEEG contact sites, where constructing the adjacency matrix is a crucial part of the GNN.
Existing studies rely on static relationships from different aspects to build adjacency matrix, e.g., correlation \cite{rf22} and distance \cite{rf23}.
Concretely, Liu et al. \cite{rf22} employ the maximum cross-correlation coefficient to define the adjacency matrix of GNN.
Dou et al. \cite{rf13} employ the results of paired t-test to construct spatial relationships between sEEG contact sites.
Daniele et al. \cite{rf23} further consider the phase relation among sEEG and utilize the phase-locking value (PLV) to construct the adjacency matrices.

However, neuroscience finds that dynamic fluctuations in neural networks are also present in the propagation of epilepsy, reflecting the evolving nature of neural interactions \cite{rf24}, \cite{rf25}.
However, the methods in the above studies lack sufficient representational capacity for epileptic information.

\section{Methods}

\subsection{sEEG Data Preprocessing}

During intracranial monitoring, sEEG data are recorded at a sampling rate of 10,000 Hz and filtered using a 50 Hz notch filter along with a 0–800 Hz lowpass filter to eliminate powerline interference and reduce noise.
We employ sEEG data collected from two conditions: one from three different behavioral states and the other under electrical stimulation (CCEP data).

\subsubsection{Preprocessing of sEEG Data from Three Behavioral States}
Different behavioral states inherently encompass  different degrees of epileptic information \cite{rf5}, \cite{rf6}, \cite{rf7}. 
To preserve the richness of epileptic information, we record the sEEG data from three behavioral states: onset, sleep and awake.
The sEEG data is downsampled to 1000 Hz and then decomposed into six frequency bands ($\delta$ band: 1-4 Hz, $\theta$ band: 4-8 Hz,  $\alpha$ band: 8-14 Hz, $\beta$ band: 14-30 Hz, low $\gamma$ band: 30-80 Hz, and high $\gamma$ band: 80-150 Hz).
We further calculate the power spectra of each frequency band by utilizing a Morlet wavelet transform with a mother wavelet of six cycles.
The absolute value of the wavelet transformation is employed as the power feature.
%
%
\subsubsection{Prepocessing of sEEG Data under Electrical Stimulation}

sEEG data under electrical stimulation, that is CCEP data, describes stable neural pathways by recording the electrical response in the brain when one cortical area is stimulated and the resulting activity is observed in another, the synaptically connected cortical region \cite{rf8}, \cite{rf9}.
CCEP stimulation is conducted in the awake state of the patient.
The electrical stimulation is performed by an electrical stimulator on each contact site.
The stimulation frequency is set to 50 Hz, and the waveform of stimulation is a square-wave pulse with intensity and duration is 5 mA and 0.3 ms, respectively.
In this study, CCEP data for each contact site, with a stimulation duration of 6 seconds, is down-sampled to 5000 Hz.
The above CCEP stimulation procedure complies with the international standard sEEG guidelines \cite{rf75}.
Besides, to minimize the influence of abnormal or pathological brain activity, the CCEP data that triggered epileptiform discharges are excluded from this study.

\subsection{Construction of sATAE-HFGCN}
In this section, we will introduce the proposed method, including the construction of sATAE, graph and HFGCN.

\subsubsection{Attention-based Autoencoder}

\begin{figure*}
	\centering
	\includegraphics[width=1\linewidth]{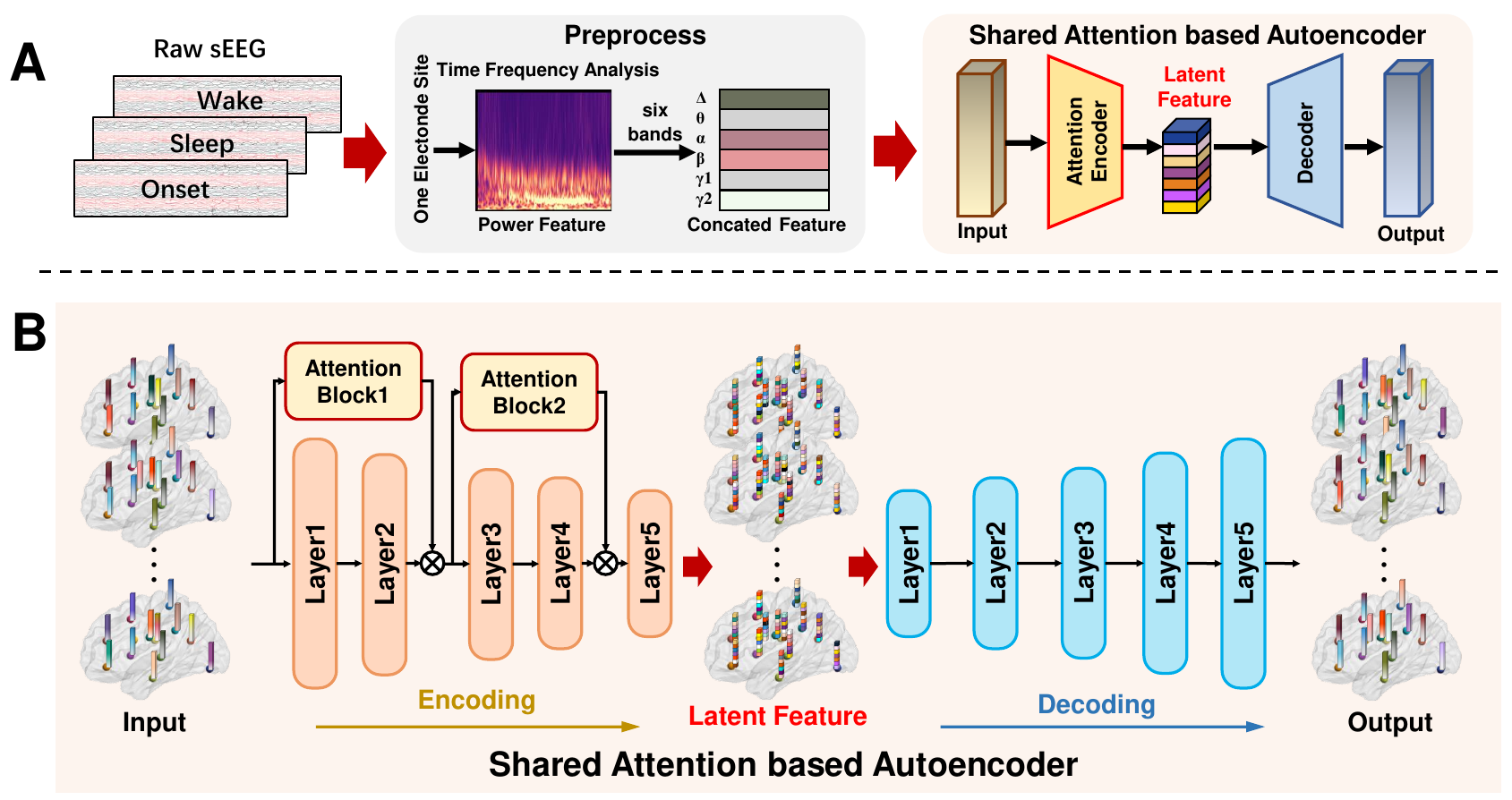}
	\caption{
		(A) sEEG processing and extraction of latent features through shared attention-based autoencdoer.
		(B) Concrete details of shared attention-based autoencoder(sATAE).}
	\label{fig:1}
\end{figure*}

In order to capture the general epileptic characteristics of all patients and enhance the quality of feature representation for epileptic information from contact sites, we propose a shared attention-based autoencoder (sATAE), whose framework is illustrated in Figure \ref{fig:1}(B).
Given that $I$ represents the dimension of the input feature, the sEEG feature from a contact site can be expressed as 
$\bold x \in \mathbb{R}^{1\times I}$.
The encoder transforms  $\bold x$  into latent representation  $\bold l$  by employing two attention blocks and five fully connected layers.
The attention block connects the input and output of every two fully connected layers.
The two fully connected layers can be expressed as:

\begin{align}
	\bold h^e_1 &= \sigma( \bold x \bold W^e_1 + \bold b^e_1), \\
	\bold h^e_2 &= \sigma(\bold h^e_1 \bold W^e_2 + \bold b^e_2),
\end{align}
where $\sigma(\cdot)$ is the $Tanh$ activation function, $\bold W^e_s$ and $\bold b^e_s$ are transformation matrix and bias in $s$-th fully connected layer ($s \in [1,2]$), respectively.
Simultaneously, input $\bold x$ also fed to attention block to generate attention coefficient. 
The output then multiplies with $\bold h^e_2$ using Hadamard multiplication to capture interdependencies between feature elements:
\begin{equation} 
	\bold h^a_{out_1} = \sigma(\text{Avgpool}(\bold x) \bold W^p_1 + \bold b^p_1) \otimes \bold h^e_2,
\end{equation}
where 
$\bold W^p_1$ and $\bold b^p_1$ is the learnable matrix and bias in the pooling path, respectively.
Avgpool($\cdot$) is average pooling operation and $\otimes$ denotes Hadamard multiplication.
Next, $\bold h^a_{out_1}$ is employed to conduct next weighting calculation (see Figure \ref{fig:1}(B)) to obtain $\bold h^a_{out_2}$, and the obtained output is then fed to a fully connected layer to obtain the latent representation $\bold l \in \mathbb{R}^{1 \times N}$, where $N$ is the length of the latent feature.
Subsequently, the decoder transforms $\bold l$ to the output $\bold y \in \mathbb{R}^{1\times L}$ by five fully connected layers.
For simplification, one of the fully connected layers in the decoder can be expressed as:
\begin{equation} 
	\bold h^d_{t} = \sigma(\bold h^d_{t-1} \bold W^d_t + \bold b^d_t),
\end{equation}
where $\bold W^d_t$ and $\bold b^d_t$ are the transformation matrix and bias of $t$-th fully connected layer in decoder ($\{t\in \mathbb{Z} \mid 1 \leq t \leq 5\}$), respectively.
We train a sATAE across all patients and minimize the mean squared error (MSE) loss between $\bold x$ and $\bold y$, the MSE loss function is expressed as:
\begin{equation}
	\mathcal{L}_{MSE}(\bold x, \bold y) = \frac{1}{I}\Vert \bold x-\bold y\Vert_2^2.
\end{equation}


\subsubsection{Construction of Graph}
In this part, we employ the latent feature from the previous part to construct the graph and provide a detailed explanation.
For each patient, the spatial positions of implanted high-density electrodes are different, resulting in different connectivity patterns. 
Therefore, we construct a graph corresponding to each patient.
The latent features of all sEEG contact sites from a patient are used as node features in a graph, denoted by $\bold X \in \mathbb R^{C \times N}$, where $C$ is the number of contact sites.
The CCEP data corresponding to a patient, $\{\bold S^{ccep}_q \in \mathbb R^{C \times T} \mid q \in [1, Q]\}$, are utilized to construct the connectivity relationships, where each $\bold S^{ccep}_q$ represents a segment of CCEP data, $Q$ is the number of CCEP stimulations a patient received, $T$ denotes the duration of each CCEP stimulation.
Concretely, the graph-structure data can be denoted as $\mathcal{G} = (\mathcal{V}, \bold A)$, where $\mathcal{V}=\{\bold{v}_1,\bold{v}_2, \dotsc, \bold{v}_C\}$ denotes the set of nodes.
$\bold{v}_i$ corresponds to the transposed features of $i$-th row of $\bold X$.
Next, the CCEP data is utilized to build an adjacency matrix  $\bold{A} \in \mathbb{R}^{C \times C}$ for $\mathcal{G}$.
In order to evaluate stable and effective connections between the stimulation and the remaining sites, statistical properties of CCEP data are employed to build connection relationships.
Paired $t$-test is able to preserve the genuine effects of electrical stimulation on neural activity while simultaneously filtering out random noise and incidental variations.
Next, we employ the paired $t$-test to evaluate the significant difference between the baseline $\bold S^{baseline}$ and CCEP data  $\bold S_q^{ccep}$.
We select 60-second stationary interictal sEEG data to establish the baseline, where interictal sEEG refers to the time period between two seizures for a patient, recording the normal and stable activities of the brain.
The interictal data is divided into ten subsegments, and the baseline $ \bold S^{baseline} \in \mathbb{R}^{C \times T}$ is obtained by averaging these subsegments.
Subsequently, the false discovery rate (FDR) is applied to 
amend the significant difference $p$-value for multiple comparisons and amended $p$-value vector denoted as $\bold p^a = [p^a_1, p^a_2, \cdots, p^a_C]^T$.
We retain contact sites that exhibit significant differences $(p^a_i<0.05, i \in [1,C])$ and obtain corresponding masks $\bold p^m = [p^m_1, p^m_2, \cdots, p^m_C]^T$, where:
\begin{equation}
	p_{i}^{m} = \begin{cases} 
		1, & \text{if } p_{i}^{a} < 0.05, \\
		0, & \text{otherwise},
	\end{cases}
\end{equation}
Subsequently, the masked CCEP data is obtained by $\bold S_q^{m \_ ccep} = \bold p^m \otimes \bold S^{ccep}_q$.
It is typically believed that the relationship between two variables is stronger when they exhibit a high correlation,  providing information about the correlation among variables.
The Pearson correlation coefficient (PPC) is a commonly used method for constructing the adjacency matrix but may result in noisy connections simultaneously.
Therefore, we modify the PPC method by setting a threshold to filter out noise connections.
The mathematical definition of PCC is given as follows:

\begin{equation}\label{eqn-3} 
	\rho_{{i},{j}} = \frac{\text{cov}(\bold v_i^{m \_ ccep}, \bold{v}_j^{m \_ ccep})}{\sigma_{\bold{v}^{m \_ ccep}_{i}}\sigma_{\bold{v}^{m \_ ccep}_{j}}},
\end{equation}
where $\text{cov}(\bold v_i^{m \_ ccep}, \bold{v}_j^{m \_ ccep})$ is covariance between $\bold v_i^{m \_ ccep}$ and $\bold v_j^{m \_ ccep}$, and $\sigma_{\bold{v}^{m \_ ccep}_{i}}$ and $\sigma_{\bold{v}^{m \_ ccep}_{j}}$ are the corresponding standard deviations.
Therefore, the element in the $i$-th row and $j$-th column of $\bold A_q$ is defined as follows:
\begin{equation}
	a_{i,j}= \begin{cases} 
		\rho_{i,j}, & \text{if } \rho_{i,j} < \rho_{\tau}, \\
		0, & \text{otherwise}.
	\end{cases}
\end{equation}
where $\rho_{\tau}$ is correlation threshold.
Then, we obtain $\{\bold A_q \in \mathbb R^{C \times C} \mid  q \in [1, Q]\}$, 
where each $\bold A_q$ represents an adjacency matrix corresponding to a CCEP data $\bold S_q^{ccep}$. 
Finally, the adjacency matrix $\bold A$ is obtained by averaging all $\bold A_q$. 
\subsubsection{Construction of Hierarchical Fusion-based Graph Convolutional Network}
The propose of HFGCN is to fuse the static and dynamic characteristics of the epileptic network hierarchically.
In HFGCN, static and dynamic graph convolutions are utilized to represent the static and dynamic aspects of epileptic network, respectively, while information from both dynamic and static networks is further fused through hierarchical weighting.
For static graph convolution, we use the Chebyshev convolution form. 
Given the adjacency matrix $\bold{A}$ of a graph $\mathcal{G}$, 
the normalized Laplacian matrix can be expressed as $ \bold{L} = \bold{I}_{n}-\bold{D}^{-\frac{1}{2}}\bold{A}\bold{D}^{-\frac{1}{2}}$, 
where $\bold{I}_{n} \in \mathbb{R}^{C\times C}$ is an identity matrix and $\bold D \in \mathbb{R}^{C \times C}$ is degree matrix.
$\bold{\widetilde{L}}$ is defined as $\bold{\widetilde{L}}= 2\bold{L}/\lambda_{max}-\bold{I}_n$, in which $\lambda_{max}$ denotes the largest eigenvalue of Laplace matrix $\bold{L}$.
Let $\Hat {\bold {H}}_{l-1}$ to be the input of $l$-th Chebyshev graph convolution layer (where the input of the first layer is $\bold X$) and $F$ is the order of Chebyshev polynomials \cite{rf33} in each layer, the corresponding output $\Hat {\bold {H}}_{l}\in \mathbb{R}^{C\times D}$ is given by:
\begin{equation}
	\Hat {\bold {H}}_{l} =  \sigma(\sum_{f     =0}^{F-1} 
	T^{f}(\widetilde{\bold{L}})\Hat {\bold {H}}_{l-1} \bold{W}_{l}^f),
\end{equation}
where $\bold{W}^{f}_{l}$ denotes a learnable transformation matrix and $T^f (\bold{\widetilde{L}})$ is the Chebyshev polynomial of order $f$ evaluated at the scaled Laplacian $\bold{\widetilde{L}}$.
Here $T^f(x)$ is computed by the stable recurrence relation $T^f(x)=2xT^{f-1}(x)-T^{f-2}(x)$, where $T^0=1$ and $T^1=x$.
For simplification, we use $\mathcal{F}(\cdot,\cdot)$ to denote graph convolution operation and $\Hat {\bold {H}}_{l}$ can be represented by:
\begin{equation}
	\Hat {\bold {H}}_{l} = \mathcal{F}(\Hat {\bold {H}}_{l-1}, \bold{W}_{l}^f),
\end{equation}
\begin{figure*}
	\centering
	\includegraphics[width=1\linewidth]{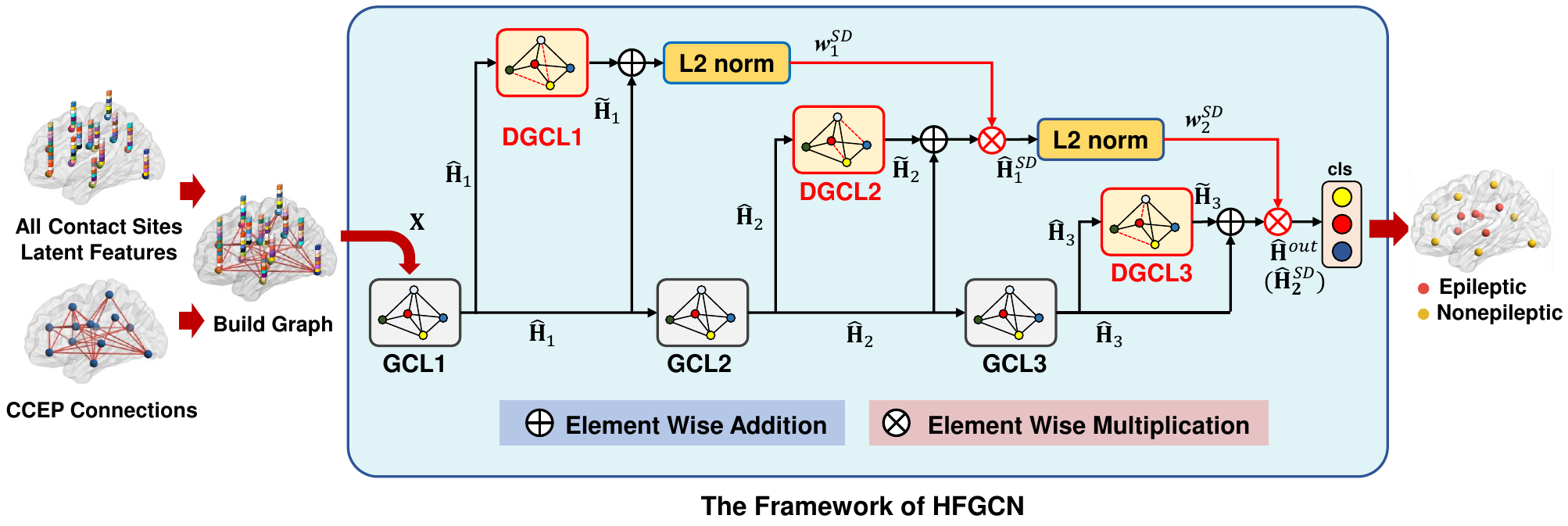}
	\caption{Construction of graph structure data and the overall framework of hierarchical fusion-based graph convolution network (HFGCN).}
	\label{fig:2}
\end{figure*}
Based on insights from neuroscience, the epileptic network is characterized by a complex interaction between dynamic and static components \cite{rf24}, \cite{rf25}.
Relying exclusively on static convolution neglects the dynamic characteristics inherent within the epileptic network.
Therefore, we incorporate dynamic graph convolution to effectively capture and supplement the dynamic information within the epileptic network.
Edge convolution (EdgeConv) \cite{rf34}  is an adaptive dynamic convolution method based on node distances, which can account for the dynamic process of neural synchronization in epilepsy.
Thus, we  employ EdgeConv to capture the dynamic characteristic.
The output of each Chebyshev graph convolution is taken as input of EdgeConv.
For $\Hat {\bold {H}}_{l}$, the $i$-th row corresponds to the features of $i$-th node, denoted by $\{\Hat{\bold {v}}^{l}_i\ \mid i \in [1, C]\}$.
Firstly, the $K$-Nearest Neighbors ($K$-NN) is employed to each node in $\Hat {\bold {H}}_{l}$.
In following, we ignore the superscript of $\Hat{\bold {v}}^{l}_i$ for simplification.
For each node $\Hat {\bold {v}_i}$, the $K$ nearest nodes $\{ \Hat{\bold v}_k^{knn} \mid k \in [1, K] \}$ are selected based on the squared Euclidean distance to $\hat{\bold {v}}_i$, 
The node pairs formed with $\Hat {\bold {v}_i}$ and its $k$-th nearest neighbor is denoted as $(\Hat {\bold {v}_i}, \Hat {\bold {v}}_k^{knn})$.
For each node pair $(\Hat {\bold {v}_i}, \Hat {\bold {v}}_k^{knn})$, a learnable approach is employed to model the connection representation:

\begin{equation}
	\bold e^{knn}_{i,k} = \Hat{\sigma}(\Hat{\sigma}(\text{concat}(\Hat{\bold v}_i, \hat{\bold v}^{knn}_k)\bold W^{knn}_1)\bold {W}_2^{knn}),
\end{equation}
where concat$(\cdot, \cdot)$ represents concatenation operation, $\Hat{\sigma}(\cdot)$ is ReLU activate function, and $\bold W^{knn}_1$, $\bold W^{knn}_2$ are transformation matrices.
Next, we update the node feature by applying aggregation operation on all connection representations with all connections emanating from $i$-th node. 
The updated $i$-th node feature $\widetilde{\bold v_i}$ can be expressed as:
\begin{equation}
	\widetilde{\bold v}_i = \text{agg}_{k=1}^{k=K}(\bold e^{knn}_{i,k}),
\end{equation}
where agg$(\cdot)$ represents aggregation function.
We apply the maximum aggregation operation and the output from EdgeConv can be expressed as $\widetilde{{\bold {H}}}_{l}$.
To integrate information captured by both static and dynamic graph convolutions, the element-wise addition is employed to $\Hat {\bold {H}}_{l}$ and $\widetilde{{\bold {H}}}_{l}$, and a node-wise $L_2$ norm is followed to generate 
weighting coefficient $\boldsymbol w_{l}^{SD}$ for the $l$-th layer:

\begin{equation}
	\boldsymbol {w}^{SD}_{l} = \Vert(\Hat {\bold {H}}_{l} + \widetilde{{\bold {H}}}_{l})\Vert_2,
\end{equation}
Subsequently, the $\boldsymbol w_{l}^{SD}$ is multiplied with the summed output obtained from the next layer of GCN and DGCN to obtain the weighted output $\Hat{\bold {H}}^{SD}_{l}$, which can be expressed as follows:

\begin{equation}
	\Hat{\bold {H}}^{SD}_{l} = 	\boldsymbol w^{SD}_{l}(\Hat {\bold {H}}_{l+1} + \widetilde{{\bold {H}}}_{l+1}),
\end{equation}
where $\Hat {\bold {H}}_{l+1}$ and $\widetilde{{\bold {H}}}_{l+1}$ represent the outputs of the $l$+1-th layer of GCN and DGCN, respectively. 
Note that the weighting process described above is not implemented for every layer in HFGCN. 
The specific structure is illustrated in Figure \ref{fig:2}.
After $L$ layers of GCN and DGCN learning and $L-1$ times of hierarchical weighting, our method fuses representations from both static and dynamic characteristics of brain networks in different hierarchies and obtain the final output of our method, denoted by $\Hat{\bold {H}}^{out}$.
Afterward, $\Hat{\bold {H}}^{out}$ is fed into a fully connected layer, and the Softmax function is applied to obtain the classification probabilities for each node.
Our method is trained by minimizing the cross-entropy loss between the predicted classification results and the true labels.

\begin{table*}
	\caption{Comparisons of the Results  (ACC(\%), Recall(\%), Precision(\%), $F_{1}$ Score(\%) and corresponding STD(\%)) of our method with different methods.}
	\label{tab:my_label1}
		\tabcolsep=0.43cm
		\renewcommand{\arraystretch}{1}
		\centering
			\normalsize
			\begin{tabular}{lccccc}
				\toprule
				
				Method & ACC / STD & Recall / STD & Precision / STD&  $F_{1}$ Score / STD \\
				\midrule
				
				
				$K$-NN($K$=10) \cite{rf35} & 68.01 / 10.78 & 46.61 / 12.33  & 48.53 / 13.26 & 47.55 / 12.79& \\
				
				
				SVM(RBF) \cite{rf36} & 64.75 / 11.06 & 51.86 / 10.74 & 50.56 / 12.38 & 51.20 / 11.54  & \\
				%
				%
				
				1D Plain CNN \cite{rf37} & 70.79 / 12.92 & 56.17 / 11.22 & 55.90 / 10.55 & 54.98 / 10.89 & \\
				
				1D ResCNN \cite{rf37} & 71.32 / 10.08 & 56.89 / 10.79 & 56.21 / 9.06 & 57.92 / 9.85& \\
				
				GCN \cite{rf38}	& 72.58 / 10.37  & 59.34 / 9.94  & 58.96 / 10.88 & 60.15 / 10.42 & \\
				
				GAT	\cite{rf41} & 72.23 / 10.06 & 60.60 / 10.71 & 58.05 / 9.23 & 59.31 / 9.94 & \\
				
				GraphSAGE \cite{rf40} & 75.56 / 9.84 & 58.15 / 9.75 & 57.33 / 10.79 & 57.74 / 10.32 & \\
				
				FastGCN \cite{rf45} & 76.62 / 9.46 & 59.30 / 9.50 & 59.94 / 8.92 & 59.62 / 9.18 & \\
				
				ResGCN+DropEdge \cite{rf44} & 76.87 / 7.11 & 64.66 / 7.36 & 63.08 / 7.44 & 63.66 / 7.40 & \\
				
				ClusterGCN	\cite{rf43} & 77.33 / 7.87 & 64.60 /  7.27 & 63.05 / 9.05 & 63.69 / 8.07 & \\
				
				ResGCN \cite{rf39}	& 75.77 / 8.25 &  63.62 / 7.04  & 62.34 / 8.77& 63.62 / 7.81 & \\				
				JK-Net \cite{rf42} & 76.12 / 7.17 & 64.33 / 6.79 & 62.96 / 7.65 &  63.69 / 7.23 & \\
				
				$\textbf{sATAE-HFGCN}$	& $\textbf{80.46 / 6.81}$ & $\textbf{67.31 / 6.54}$ & $\textbf{66.04 / 6.47
				}$ & $\textbf{66.67 / 6.52}$ & \\
				\bottomrule
				
			\end{tabular}
		\renewcommand{\rmdefault}{ptm} 
	\end{table*}
\section{Experiments}
In this section, we conduct comprehensive experiments to evaluate the performance of our method for sEEG SOZ identification.
\subsection{Dataset}
This study was approved by the Ethics Committee of Fujian Medical University Affiliated Union Hospital  (Title: The SINO-robot and its vascular rendering visualization imaging technology facilitate sEEG procedures for treating drug-resistant epilepsy; ethics approval number: 2021KY119). 
The analysis included data from patients with refractory temporal epilepsy who received neurosurgical treatment at Fujian Medical University Affiliated Union Hospital.
For each patient, a diagnosis of drug-resistant epilepsy led to a recommendation for surgical intervention.
Before surgery, non-invasive evaluations are performed, including the collection of clinical history, magnetic resonance imaging (MRI) and positron emission tomography (PET) scans to assess habitual seizures, and scalp-EEG recordings to capture ictal EEG activity.
Subsequently, high-density electrodes implantation surgery is conducted.
Each patient underwent over a week of invasive monitoring with implanted depth electrodes.
SOZ is preliminarily determined by integrating results from brain MRI/PET, regional metabolism, CCEP stimulation and relevant medical assessments.
Subsequently, sEEG-guided radiofrequency thermocoagulation (RFT) to disrupt the  preliminarily identified SOZ.
After RFT surgery, if the patient remains seizure freedom over a 2-3 year follow-up period, the RFT area is designated as the final effective SOZ label.
Our dataset consists of patients who meet the aforementioned medical assessment criteria.

Our dataset records sEEG of 17 patients with drug-resistant temporal lobe epilepsy.
sEEG data is collected by using depth electrodes stereotactically implanted into the patients' temporal structures with the SINO-robot (Sinovation depth electrode, Beijing, China).
For each patient, 9 to 19 high-density electrodes are implanted, and each high-density electrode has a total of 8 to 16 contact sites (length: 2mm, diameter: 0.8mm).
Thus, each patient has a total of 127 to 254 contact sites for sEEG recording.
Among these, 23 to 68 contact sites located within the SOZ. 
The labels for contact sites are determined collaboratively by eight epilepsy neurosurgeons.
For sEEG data from three behavior states, each state of data contains 60-seconds duration.
Each segment of CCEP data is a 6-second recording of electrical stimulation from medial temporal lobe.
\subsection{Implementation Details}
For sATAE, batch size and number of epochs are set to 16 and 30, respectively.
Adam optimizer is used for optimization with a learning rate of 0.002.
For HFGCN, the correlation threshold $\rho_\tau$ is set to 0.3.
The number of static and dynamic graph convolution layers is set to 3.
The Chebyshev kernel size $F$ in the static graph convolution is set to 3, and $K$ in the dynamic graph convolution is set to 10. 
Optimization is performed using the Adam optimizer with a learning rate of 0.005, and the model is trained for 150 epochs.
Our model is implemented with PyTorch trained on a Geforce RTX 4090 GPU.
sEEG data from a single patient's three behavior states is employed to build node features, and CCEP data is applied to generate the corresponding adjacency matrix. 
We follow the experimental protocol for node classification task established in benchmark graph datasets \cite{rf76}.
Specifically, for each patient, we utilize 10\% of the latent features for training, 20\% for validation and the remaining 70\% for testing.
The results are obtained separately on each patient's testing set.
We use averaged Accuracy (ACC), Recall, Precision, and $F_1$ Score and their corresponding standard deviation (STD) across all patients to evaluate the performance.
SOZ contact sites are labeled as positive and non-SOZ contact sites are labeled as negative.

\begin{table*}
	\caption{Comparisons of the results (ACC(\%), Recall(\%), Precision(\%), $F_{1}$ Score(\%) and corresponding STD(\%)) of our method on different frequency bands.}
	\label{tab:my_label2}
		\tabcolsep=0.47cm
		\normalsize
		\centering
		\renewcommand{\arraystretch}{1.0}
			\begin{tabular}{lccccc}
				\toprule
				Frequency Band & ACC / STD & Recall / STD & Precision / STD &  $F_{1}$ Score / STD \\
				\midrule
				
				$\delta$ band & 78.82 / 6.29
				& 64.82 / 6.34 & 63.67 / 6.90 & 63.24 / 6.60 & \\
				
				
				$\theta$ band & 78.16 / 7.38 & 62.93 / 7.66& 61.25 / 7.83& 62.09 / 7.72 & \\

				$\alpha$ band & 76.97 / 6.74 & 63.46 / 7.10& 60.60 / 6.76& 61.99 / 6.93 & \\
				
				$\beta$ band & 75.24 / 7.01 & 62.90 / 6.95 & 61.01 / 7.38 & 61.95 / 7.14 & \\
				
				low $\gamma$ band & 77.88 / 6.18 & 65.32 / 6.37 & 64.30 / 6.43 & 64.81 / 6.39 & \\
				
				high $\gamma$ band	& 78.14 / 6.87 & 65.15 / 6.89 & 63.99 / 7.07& 64.59 / 6.98 & \\
				
				all bands 
				& $\textbf{80.46 / 6.81}$ & $\textbf{67.31 / 6.54}$ & $\textbf{66.04 / 6.47
				}$ & $\textbf{66.67 / 6.52}$ & \\
				\bottomrule
				
			\end{tabular}
	\end{table*}
	\begin{table*}[h]
		\caption{Comparisons of the results (ACC(\%), Recall(\%), Precision(\%), $F_{1}$ Score(\%) and corresponding STD(\%)) of our method with CCEP data from different anatomical structures.}
		\label{tab:my_label3}
		\centering
		\tabcolsep=0.4cm
			\normalsize
			\renewcommand{\arraystretch}{1}
			
			\begin{tabular}{cccccc}
				\toprule[1pt]
				Anatomical Structure & ACC / STD & Recall / STD & Precision / STD &  $F_{1}$ Score / STD \\
				\midrule
				
				AMY  & $\textbf{79.27 / 7.10}$ & $\textbf{65.03 / 6.88}$ & $\textbf{64.87 / 7.33}$ &  $\textbf{64.95 / 6.75}$ & \\

				HIP & 78.98 / 8.24 & 64.88 / 7.04 & 63.31 / 7.90 & 64.09 / 7.43 & \\
				
				OC & 77.97 / 7.15 & 63.46 / 7.53 & 60.60 / 8.80 & 61.98 / 8.11 & \\
				
				PHG & 78.24 / 8.89 & 64.50 / 8.76 & 62.01 / 9.28 & 63.24 / 9.02 \\
				
				\bottomrule[1pt]
				
			\end{tabular}
		\renewcommand{\rmdefault}{ptm} 
	\end{table*}
\subsection{Performance Comparison of SOZ Classification with Different Methods}
We select two types of methods for performance comparison with our method: traditional machine learning method and deep learning method.
Two traditional machine learning methods include $K$-NN and support vector machine (SVM).
Ten deep learning methods are selected from two perspectives:  non-spatial learning method and spatial learning method.
Two non-spatial learning method include 1D Plain CNN and 1D ResCNN.
Eight spatial learning method include two baseline methods (GCN and graph attention network (GAT)), two node sampling methods (GraphSAGE, FastGCN), two edge dropping methods (ResGCN+DropEdge and ClusterGCN) and two skip connection methods (ResGCN and JK-Net).

For each patient, the latent features from the six frequency bands across the three behavioral states are concatenated to form the feature of each contact site.
We compare the performance of our method with all selected methods and show the results in Table \ref{tab:my_label1}.
By incorporating the use of sATAE and HFGCN,
our method achieves the best performance across all four metrics
(ACC 80.46\%, Recall 67.31\%, Precision 66.04\%, $F_1$ score 66.88\%), surpassing the second best method by 3.13\% in ACC, 2.65\% in Recall, 2.99\% in Precision, and 2.98\% in $F_1$ Score.
We consistently find that deep learning methods are superior to traditional machine learning methods, which means that the features learned by deep learning methods provide better representational capability for sEEG SOZ identification.
Additionally, within deep learning methods, graph-based methods perform better than CNN-based methods,
suggesting that graph-based methods are more effective in leveraging spatial information for sEEG SOZ identification.

\subsection{Performance Comparison of SOZ Classification across Different Frequency Bands}
Different frequency bands of sEEG exhibit distinct characteristics of epileptic information.
Therefore, we investigate the performance of our method by employing latent features from different frequency bands for sEEG SOZ identification.
For the graph corresponding to each patient, the latent feature from single frequency band across the three behavioral states are concatenated to form the node feature.

	The identification results of the six frequency bands are given in Table \ref{tab:my_label2}.
	Our method achieves the best performance in the $\delta$ band (ACC 78.82\%, Recall 64.82\%, Precision 63.67\%, $F_1$ score 63.24\%) when a single frequency band is used as input.
	It can be inferred that the $\delta$ band contains more SOZ-related information associated with abnormal neuronal and pathological activity, this is consistent with previous finding \cite{rf48}.
	It is worth noting that the results for the high-frequency bands (low $\gamma$ and high $\gamma$ bands) achieve comparable performance to those on the low-frequency bands ($\delta$\ and $\theta$ bands).
	This phenomenon can be explained by the finding of cross-frequency coupling, where strong neuronal synchrony between low frequency and high frequency oscillations \cite{rf50}.
	Nevertheless, the mid-frequency components still carry some valuable epileptic information. 
	This is demonstrated by the results, where our method achieves the highest accuracy when all frequency bands are utilized as input.
		\begin{table*}[h]
	\caption{Comparisons of the results (ACC(\%), Recall(\%), Precision(\%), $F_{1}$ Score(\%) and corresponding STD(\%)) of our method with different behavioral states.
	}
	\label{tab:my_label6}
	\centering
	\renewcommand{\arraystretch}{1}
	\tabcolsep=0.38cm
	\normalsize
		\begin{tabular}{lcccccc}
			\toprule
			Behavior state  & ACC / STD & Recall / STD & Precsion / STD & $F_1$ Score / STD\\
			\midrule
			
			Wake & 74.20 / 8.29 & 62.56 / 7.34 & 62.39 / 8.95
			& 62.47 / 8.05 \\ 
			
			Sleep & 76.93 / 7.54 & 64.13 / 7.19 & 63.24 / 7.64
			& 63.68 / 7.42 \\
			
			Seizure  & 76.76 / 7.97 & 64.34 / 6.85 & 63.76 / 7.50 & 64.05 / 7.14 & \\
			
			Wake+Sleep & 77.32 / 7.13 & 65.92 / 6.21 & 64.31 / 6.60& 65.11 / 6.39\\
			
			Wake+Seizure & 78.76 / 6.74 & 65.78 / 7.03 & 65.29 / 7.22 & 65.53 / 7.13 \\
			
			Sleep+Seizure & 78.94 / 6.15& 65.81 / 6.43 & 64.87 / 5.96 & 65.36 / 6.17 \\

			All States & $\textbf{80.46 / 6.81}$ & $\textbf{67.31 / 6.54}$ & $\textbf{66.04 / 6.47
			}$ & $\textbf{66.67 / 6.52}$ & \\

			\bottomrule
		\end{tabular}
\end{table*}
\begin{table*}[h]
	\caption{The results (ACC(\%), Recall(\%), Precision(\%), $F_{1}$ Score(\%) and corresponding STD(\%)) of module analysis for our method.}
	\label{tab:my_label4}
	\tabcolsep=0.33cm
	\renewcommand{\arraystretch}{1}
	\centering
	\normalsize
		\begin{tabular}{lcccccc}
			\toprule
			Method  & ACC / STD & Recall / STD & Precsion / STD & $F_1$ Score / STD\\
			\midrule
			
			w/ AutoE & 67.84 / 8.13 & 50.00 / 7.28& 50.81 / 6.99
			& 50.40 / 7.11 \\ 
			
			w/ sATAE & 69.62 / 7.66 & 57.75 / 6.35 & 58.14 / 7.48
			& 57.94 / 6.87 \\
			
			w/ sATAE+GCN & 74.58 / 7.79 & 63.34 / 7.48 & 60.96 / 7.83 & 62.13 / 7.64 & \\
			
			w/ sATAE+DGCN & 76.52 / 7.64 & 64.78 / 7.25& 64.31 / 7.68 & 64.54 / 7.44 \\
			
			Full model  & $\textbf{80.46 / 6.81}$ & $\textbf{67.31 / 6.54}$ & $\textbf{66.04 / 6.47
			}$ & $\textbf{66.67 / 6.52}$ & \\
			
			\bottomrule
		\end{tabular}
\end{table*}
	\subsection{Performance Comparison of SOZ Classification with CCEP Data from Different Medial Temporal Lobe Regions}
	
	In the previous parts, we build adjacency matrix $\bold A$ by averaging the connectivity relationships generated by $Q$ times CCEP stimulation in the medial temporal lobe.
	The medial temporal lobe can be more precisely subdivided into four anatomical structures, including the amygdala (AMY), hippocampus (HIP), olfactory cortex (OC) and parahippocampal gyrus (PHG).
	Meanwhile, CCEP data and the number of stimulations vary across different anatomical structures.
	
	Here, we employ CCEP data from different anatomical structures to generate the corresponding adjacency matrix $\bold A^{region} \in \{\bold {A}^{AMY}, \bold {A}^{HIP}\, \bold {A}^{OC}, \bold {A}^{PHG}\}$ for SOZ identification, and the results are shown in Table \ref{tab:my_label3}.
	Our method, using CCEP data from the AMY, achieves the best performance with an ACC of 79.27\%, Recall of 65.03\%, Precision of 64.87\%, and $F_{1}$ Score of 64.95\%.
	Note that using CCEP data from the HIP can achieve a performance comparable to that of using CCEP data from the AMY.
	It can be inferred that the AMY and HIP represent the hotspots for initiating epileptic activities \cite{rf13}, \cite{rf51}.
	However, the results for CCEP data from the OC and 
	PHG underperform those from the AMY and HIP.
	This can be demonstrated by the fact that the OC and PHG yield to passive activation during epileptic activities  \cite{rf52}, \cite{rf53}.
	\subsection{Performance Comparison of SOZ Classification across Different Behavior States}
	According to studies in neuroscience \cite{rf5}, \cite{rf6}, \cite{rf7}, different behavior states contain diverse seizure-related information.
	Therefore, we conduct the experiment on our method by using sEEG data from different combinations of behavioral states for the sEEG SOZ identification.
	For the graph corresponding to each patient, the latent features from all frequency band across different combinations of behavioral states are concatenated to form the node feature.
	
	The results are listed in Table \ref{tab:my_label6}.
	We observe that the result of sleep and seizure exhibit comparable performance when using a single behavioral state as input.
	Specifically, the ACC of seizure (76.76\%) is slightly lower than that of sleep (76.93\%), but both Recall (64.34\%), Precision (63.76\%) and $F_1$ Score (64.05 \%) are marginally higher for seizure compared to sleep (Recall 64.13 \%, Precision 63.24\%, $F_1$ Score (63.68 \%).
	Furthermore, the combination of sleep and seizure (sleep+seizure) achieves the best performance compared to the other two combinations (wake+sleep, wake+seizure) when two behavioral states are used as inputs.
	The described findings can be explained by the presence of seizure-like discharges during the sleep state, which contributes to sEEG SOZ identification \cite{rf54}, \cite{rf55}.
	Nevertheless, the awake state still contains valuable seizure-related information \cite{rf56}, as evidenced by the method's best performance when all behavioral states are used as input.
		\begin{table*}[h]
		\caption{The results (ACC(\%), Recall(\%), Precision(\%), $F_{1}$ Score(\%) and corresponding STD(\%)) of different fusion strategies in HFGCN.}
		\label{tab:my_label10}
		\tabcolsep=0.38cm
		\renewcommand{\arraystretch}{1}
		\centering
		\normalsize
			\begin{tabular}{lcccccc}
				\toprule
				Model  & ACC / STD & Recall / STD & Precsion / STD & $F_1$ Score / STD\\
				\midrule
				
				w/ sATAE+FusionF1 & 76.95 / 8.02 & 65.24 / 7.51 & 65.08 / 7.91 & 65.16 / 7.71 \\
				
				w/ sATAE+FusionF2 & 78.34 / 7.58 & 65.52 / 7.40 & 66.37 / 7.73 & 65.94 / 7.56 \\
				
				Full model  & $\textbf{80.46 / 6.81}$ & $\textbf{67.31 / 6.54}$ & $\textbf{66.04 / 6.47
				}$ & $\textbf{66.67 / 6.52}$ & \\
				
				\bottomrule
			\end{tabular}
	\end{table*}
	\begin{table*}[h]
		\caption{The Results (ACC(\%), Recall(\%), Precision(\%), $F_{1}$ Score(\%) and corresponding STD(\%)) of attention blocks implement in different modules.}
		\label{tab:my_label5}
		\tabcolsep=0.45 cm
		\centering
		\renewcommand{\arraystretch}{1.0}
			\normalsize
			\begin{tabular}{lcccccccc}
				\toprule
				Model  & ACC / STD & Recall / STD & Precsion / STD & $F_1$ Score / STD\\
				\midrule
				
				w/ AutoE & 79.33 / 7.33 & 62.75 / 6.94 & 61.38 / 6.65
				& 62.06 / 6.81  \\ 
				
				w/ sATAE(E)  & $\textbf{80.46 / 6.81}$ & $\textbf{67.31 / 6.54}$ & $\textbf{66.04 / 6.47
				}$ & $\textbf{66.67 / 6.52}$ & \\ 
				
				w/ sATAE(D) & 78.78 / 8.85& 62.13 / 7.95 & 61.66 / 7.54& 61.89 / 7.72 & \\
				
				w/ sATAE(ED) & 79.52 / 7.11 & 64.78 / 7.62 & 64.31 / 7.72 & 64.54 / 7.64 \\   
				\bottomrule
			\end{tabular}
	\end{table*}
	\subsection{Ablation Study of sATAE-HFGCN}
	In this section, we conduct ablation studies of our proposed method, including module analysis,  fusion strategy analysis and parameter analysis.
	
	\subsubsection{Module Analysis in sATAE-HFGCN}
	To analyze the contribution of different modules in the proposed method comprehensively, we sequentially incorporate the proposed modules based on the autoencoder. 
	We show the results in Table \ref{tab:my_label4}, in which w/ AutoE and w/ sATAE represent autoencoder and attention-based autoencoder, respectively.
	w/ sATAE+GCN and w/ sATAE+DGCN represent sATAE with GCN and DGCN, respectively.
	
	Compared to AutoE, w/ sATAE shows better results, indicating that the introduction of the attention block into the autoencoder effectively captures interdependencies between feature elements in contact sites for epileptic information, thereby enhancing the representational capability of sEEG SOZ identification.
	Comparing w/ sATAE+GCN, w/ sATAE+DGCN and AutoE, w/ sATAE, 
	we can observe a significant improvement in the method performance by introducing graph learning after the pre-trained sATAE.
	This result demonstrates that graph-based method effectively learn the spatial characteristics of the epileptic network.
	Compare to the full model,  both w/ sATAE+GCN and w/ sATAE+DGCN exhibit poorer performance, implying that comprehensively considering both static and dynamic components of epileptic network can enhance performance for sEEG SOZ idenification.
	\subsubsection{The Analysis of Fusion Strategy in HFGCN}
	
\begin{figure}
		\centering
		\includegraphics[width=1\linewidth]{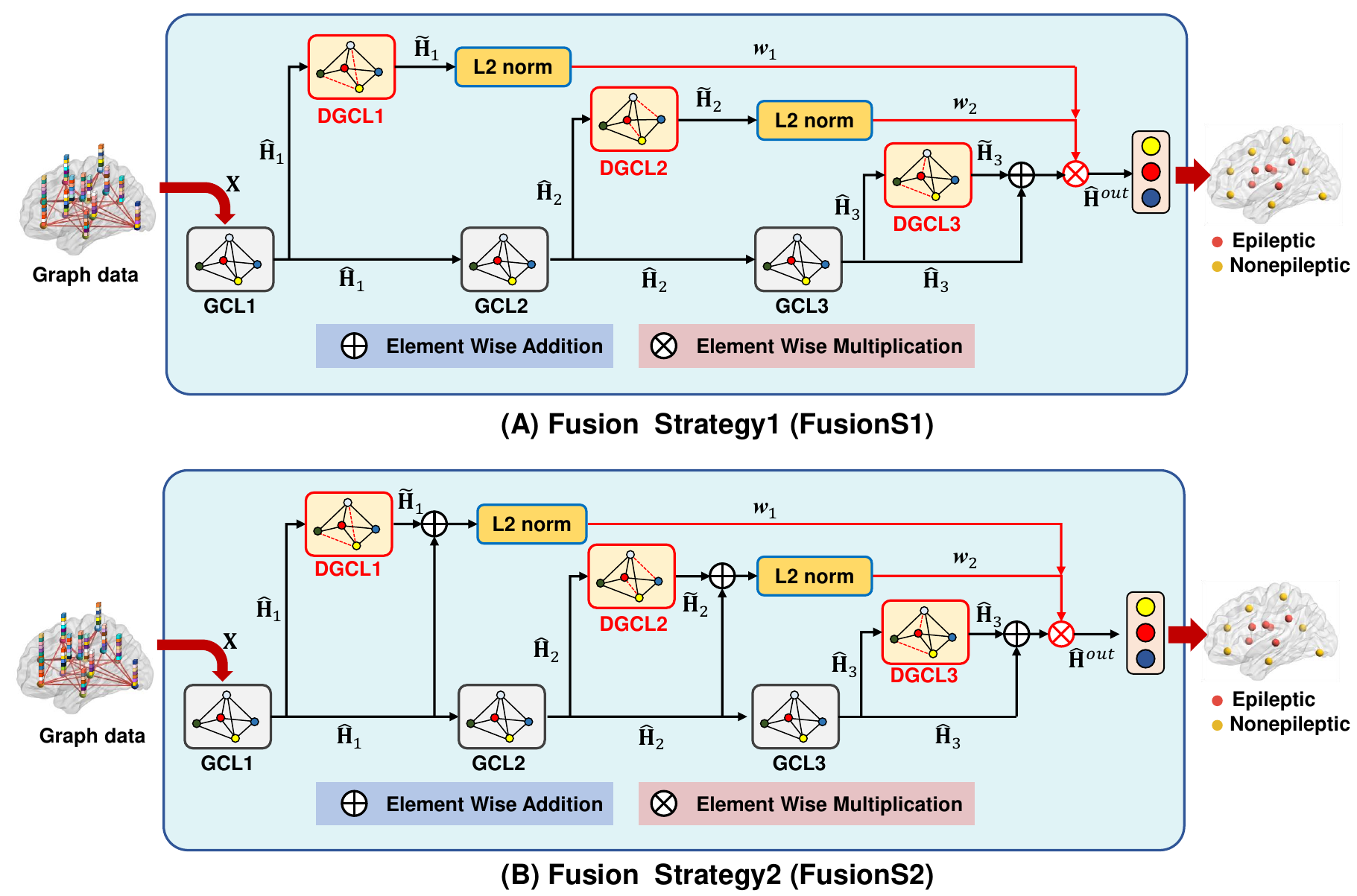}
		\caption{Two Fusion Strategies in HFGCN for sEEG SOZ Identification.}
		\label{fig:5}
	\end{figure}

	In HFGCN, the fusion strategy is implemented through two operations: hierarchical weighting and feature summation.
	In order to analyze the impact of these two operations, we simplify fusion strategy of HFGCN into two fusion strategies, referred to as FusionS1 and FusionS2, as shown in Figure \ref{fig:5}(A) and Figure \ref{fig:5}(B), respectively.
	FusionS1 can be regarded as removing both hierarchical weighting and feature summation from HFGCN, applying the generated weights directly to the final output.
	FusionS2 can be viewed as introducing feature summation based on FusionS1, while the full model can be viewed as introducing hierarchical weighting based on FusionS2.
	The results are shown in Table \ref{tab:my_label10}, where sATAE+FusionF1 and sATAE+FusionF1 represent sATAE with FusionF1 and FusionF2, respectively.
	
	Compared to w/sATAE+FusionF1, w/sATAE+FusionF2 achieves better results.
	This improvement can be attributed to the  feature summation of the corresponding static and dynamic convolutions, which enriches the feature information.
	When comparing w/sATAE+FusionF1 to the full model, the full model shows superior performance, indicating that the hierarchical weighting effectively captures information interactions in hierarchical levels across static and dynamic graph convolutions.
	\subsubsection{Effects of Attention Block in sATAE}
	
	The attention block implement in the encoder, the decoder, or both in sATAE may lead to distinct performances.
	Therefore, we conduct experiments by implementing the attention block in different modules.
	The results are shown in Table \ref{tab:my_label5}.
	In the table, w/ AutoE denotes autoencoder. w/ sATAE(E), w/ sATAE(D) and w/ sATAE(ED) represent attention blocks being applied in the encoder, decoder, and both encoder and decoder, respectively. 
	We ignore the HFGCN for simplification.
	
	Compared to AutoE, w/ sATAE achieves a significant improvement.  
	This result demonstrates that the proposed attention block effectively captures the district feature interdependencies between feature elements and improves the performance. 
	Next, we replace the pooling operation with unpooling in the attention block and introduce the attention block into the decoder to evaluate its performance.
	Comparing w/ sATAE(D) and w/ sATAE(E), moving the attention blocks to the decoder lead to performance degradation.
	This indicates that the attention block enhances feature extraction and contextual representation in the encoder, but it may disrupt the intended feature reconstruction process in the decoder.
	Note that the autoencoder structure is typically symmetric in related studies.
	Therefore, we introduce the attention block into both the encoder and decoder to construct a symmetric sATAE (sATAE(ED)).
	Compared to w/ sATAE(D), w/ sATAE(ED) demonstrates improved performance, further validating the critical role of the attention mechanism in the encoder.
	\subsubsection{Parameter Analysis}
\begin{figure}
	\centering
	\includegraphics[width=0.9\linewidth]{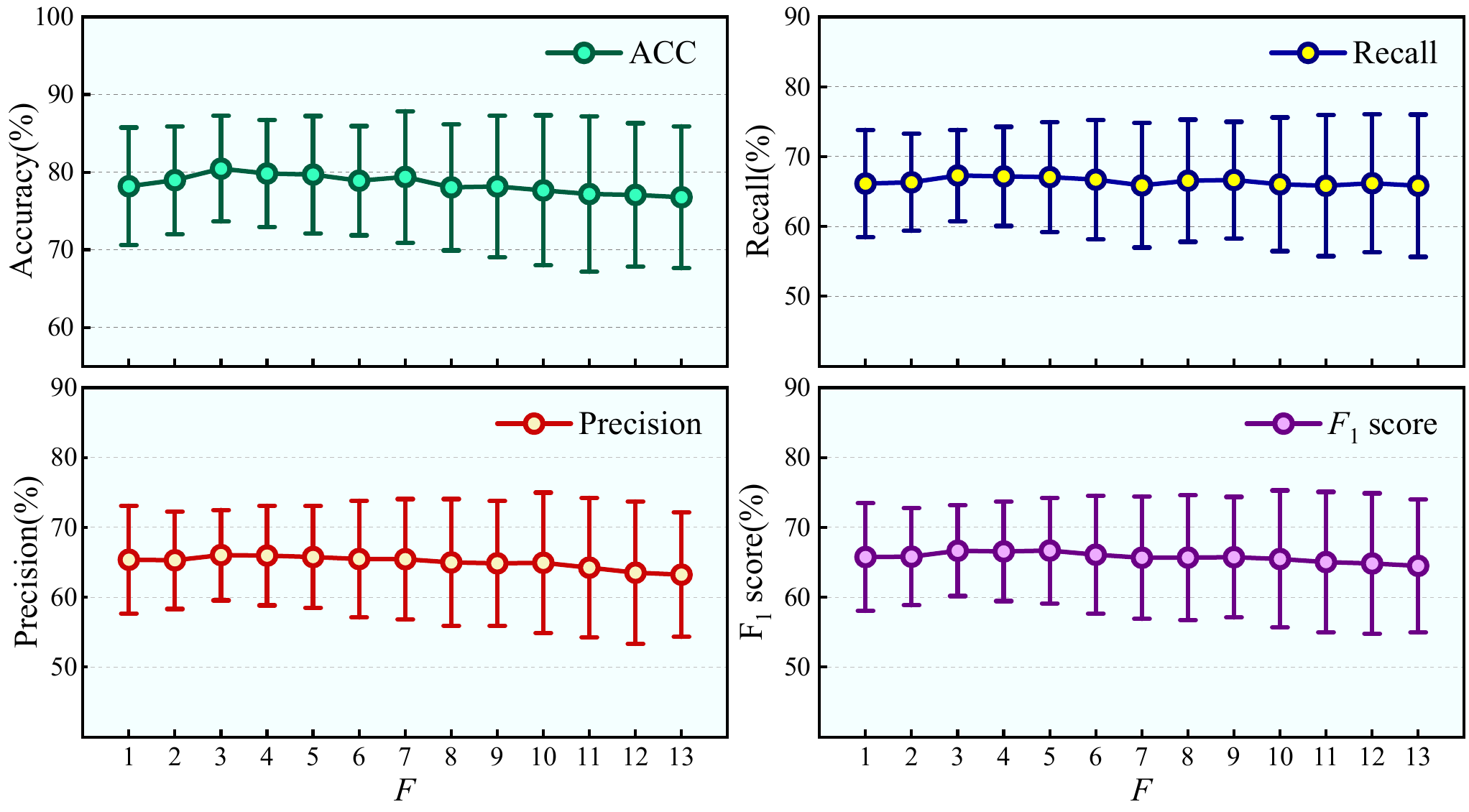}
	\caption{The method performance with different Chebyshev filter size $F$ for sEEG SOZ idenification.}
	\label{fig:4}
	\end{figure}

	Chebyshev filter size $F$ in GCN and $K$ value in DGCN are key hyperparameters in our method.
	Selecting different values of these two hyperparameters will significantly impacts the performance for sEEG SOZ identification.
	Thus, in this part, we conduct experiments to analyze the effects of different hyperparameters on sEEG SOZ identification.
	First, we set Chebyshev filter sizes $F$ from 1 to 10 separately. 
	The results are shown in Figure \ref{fig:4}. 
	It is obvious that sATAE-HFGCN achieves the best performance when $F$=3.
	Subsequently, the performance of our method shows a relatively downward trend.
	We infer that the performance decline is attributable to the effects of over-smoothing.
	
\begin{figure}
	\centering
	\includegraphics[width=0.9\linewidth]{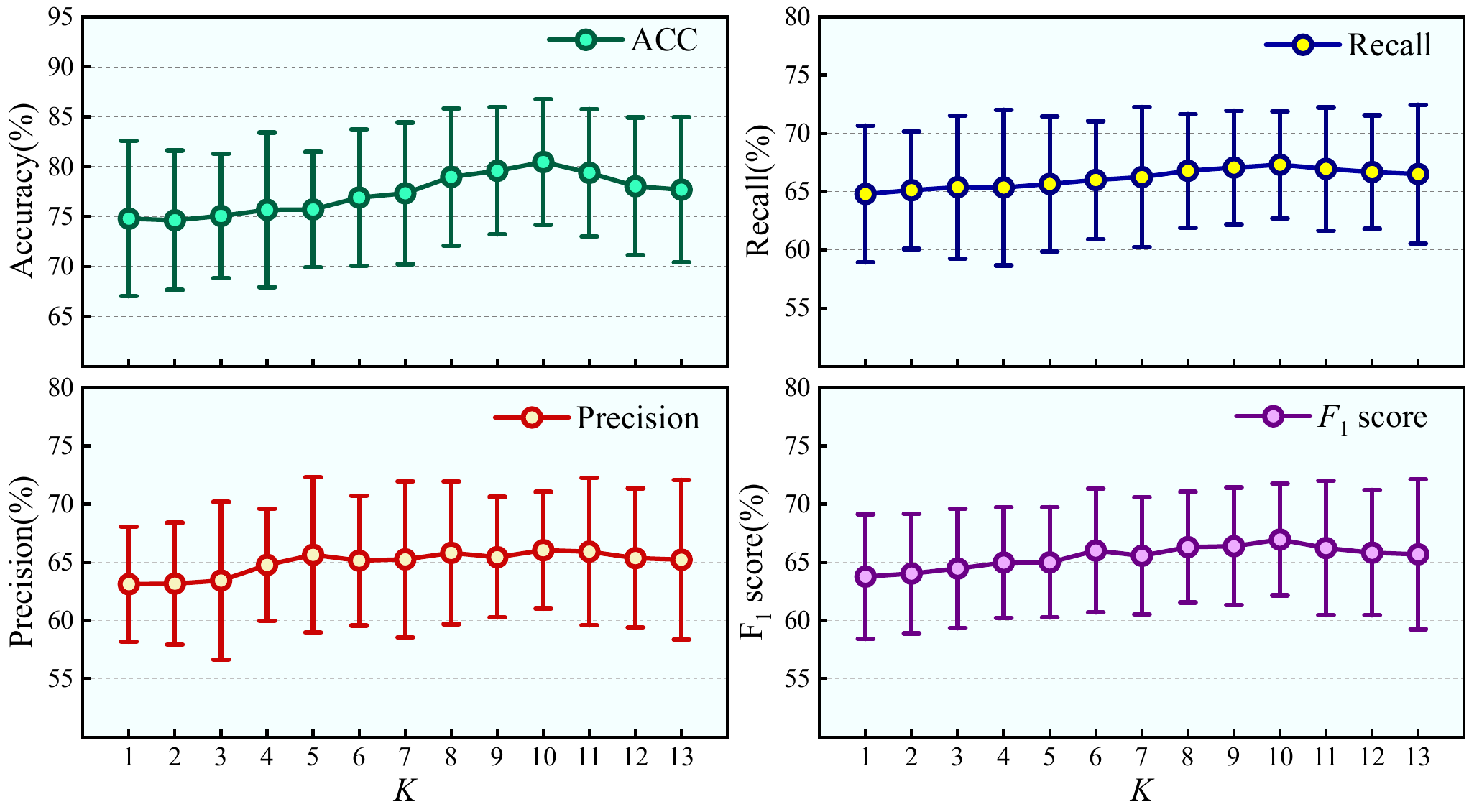}
	\caption{The method performance with different $K$ value in DGCN for sEEG SOZ idenification.}
	\label{fig:3}
	\end{figure}

	In the DGCN, the selection of $K$ value in $K$-NN is empirical.
	Here, we explore the performance of our method with different $K$ values in DGCN. 
	We set $K$ from 1 to 13 and conduct experiments on all patients, the results are presented in Figure \ref{fig:3}.
	We can see that, 
	the performance gradually improves
	and achieves the best performance when $K$ is set to 10.
	Then the performance shows a slow downward trend when $K$ is greater than 10.
	This may be because increasing the number of $K$ incorporates more nodes into feature integration, which reduces the discriminative capability of the features.
	
	
\section{Conclusion}
In this work, we propose a novel autoencoder and graph-based learning method for sEEG SOZ identification.
We design the sATAE, which effectively enhances the generalizability in representing epileptic information. 
The introduced attention block in sATAE effectively mitigate the insufficient interdependencies between feature elements in contact sites.
Moreover, the HFGCN further integrate both static and dynamic characteristics from the epileptic network by utilizing hierarchical weighting across different perspectives.
Comprehensive experiments are performed on self-build dataset comprising 17 patients with temporal lobe epilepsy.
The results demonstrate promising performance and validate the effectiveness of the proposed method.  
Our method provides a novel solution for sEEG SOZ identification.

\section*{CRediT Authorship Contribution Statement}
\noindent 
\textbf {Huachao Yan:} Writing - original draft, Conceptualization, Methodology. 
\textbf {Kailing Guo:} Writing - review\&editing, Methodology, Project Administration. 
\textbf {Shiwei Song:} Investigation, Project Administration. 
\textbf {Yihai Dai:} Data Curation.
\textbf {Xiaoqiang Wei:} Data Curation. 
\textbf{Xiaofen Xing:} Supervision, Resources.
\textbf{Xiangmin Xu:} Supervision, Funding Acquisition.

\section*{Declarations}
\noindent The authors declare that there is no conflict of interest in this work.

\section*{References}
\bibliographystyle{IEEEtran}
\bibliography{cas-refs}


\end{document}